\documentclass{Interspeech2024}

\interspeechcameraready

\usepackage{cite}  
\usepackage{multirow}
\usepackage{tablefootnote}
\graphicspath{ {./figures/} }

\usepackage{tikz}
\usepackage{everypage}

\newcommand\BackgroundText{%
  \begin{tikzpicture}[remember picture,overlay]
    \node [rotate=0, scale=1.3, text opacity=1.0, color=gray] at (current page.south west) [anchor=south west, xshift=0.7cm, yshift=1.9cm] {accepted at Interspeech 2024};
  \end{tikzpicture}
}

\AddEverypageHook{\BackgroundText}

\title{Meta Learning Text-to-Speech Synthesis in over 7000 Languages}

\name[affiliation={1}]{Florian}{Lux}
\name[affiliation={1}]{Sarina}{Meyer}
\name[affiliation={2}]{Lyonel}{Behringer}
\name[affiliation={2}]{Frank}{Zalkow}
\name[affiliation={3}]{Phat}{Do}
\name[affiliation={3}]{\\Matt}{Coler} 
\name[affiliation={2}]{Emanuël A. P.}{Habets}
\name[affiliation={1}]{Ngoc Thang}{Vu}

\address{
  $^1$University of Stuttgart, Germany\\
  $^2$Fraunhofer IIS Erlangen, Germany \\
  $^3$University of Groningen, The Netherlands}

\email{florian.lux@ims.uni-stuttgart.de}

\begin{document}

\maketitle 

\begin{abstract}
In this work, we take on the challenging task of building a single text-to-speech synthesis system that is capable of generating speech in over 7000 languages, many of which lack sufficient data for traditional TTS development. By leveraging a novel integration of massively multilingual pretraining and meta learning to approximate language representations, our approach enables zero-shot speech synthesis in languages without any available data. We validate our system's performance through objective measures and human evaluation across a diverse linguistic landscape. By releasing our code and models publicly, we aim to empower communities with limited linguistic resources and foster further innovation in the field of speech technology.
\end{abstract}

\section{Introduction}
The field of text-to-speech (TTS) synthesis offers a crucial component across a variety of applications and research fields, including accessibility features for the visually impaired, medical applications, language learning tools, language revitalization, voice privacy, literary studies, personal assistants, and entertainment. However, out of the over 7000 languages in the world\footnote{According to Glottolog: \url{http://glottolog.org}}, only a few communities currently have access to a high-quality, controllable TTS system in their native language. 

Prior work on massively multilingual TTS (i.e., dealing with hundreds of languages) is sparse. The MMS models \cite{pratap2023scaling} cover 1107 languages by combining self-supervised pretraining with supervised finetuning, resulting in one single-speaker monolingual model per language with remarkable quality. Similarly, the authors of the 
CMU Wilderness dataset \cite{black2019wilderness} train 699 monolingual models in a fully supervised manner. Virtuoso \cite{saeki2023virtuoso} is a 101-language model trained in a semi-supervised manner that does not need paired data, but still requires unpaired adaptation data.
Other works on multilingual and low-resource TTS, while operating on a smaller scale, explore transfer learning \cite{tu2019endtoend, 3403331}, dual transformation \cite{3403331}, meta learning \cite{lux2022laml, lux2022lrms}, or separating the semantic level from the acoustic level \cite{kharitonov2023speartts}. Mismatches in phoneme sets are handled by employing specialized representations as the input, such as bytes \cite{li2019bytes} or linguistically-motivated features \cite{lux2022laml, do2023strategies}.

\begin{figure}[t]
    \centering
    \includegraphics[trim={0 .3cm 0 .5cm},clip, width=.5\textwidth]{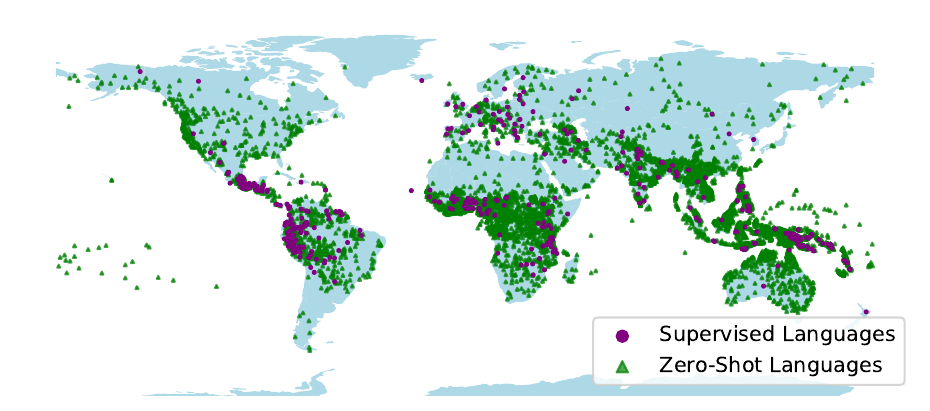}
    \caption{An overview of the coverage of supervised (462) and zero-shot (6750) languages in our work on the world map. \vspace{-1em}}
    \label{fig:worldmap}
\end{figure}

In this work, we present the first TTS system that can synthesize speech in a total of 7212 languages, covering nearly all spoken languages cataloged in Glottolog~\cite{hammarstrom2015glottolog}. We achieve this by pretraining a TTS model on a massive scale of 462 languages with a total of over 18,000 hours of paired data, which we collected from publicly available sources. The underlying TTS model is designed to be language agnostic except for a language embedding, which is used as a conditioning signal. While collecting this data and training such a model is already challenging, the resulting system still covers less than 6.4\,\% of all considered languages, illustrated in Figure~\ref{fig:worldmap}. For the remainder, we leverage the embeddings of supervised languages to approximate those of unseen languages, sharing an otherwise language-agnostic model across all languages. To achieve this, we make use of meta learning under the learn-to-compare framework, similar to Siamese nets \cite{bromley1993signature}. Using these predicted language embeddings during inference, our system can generate speech even for unseen languages. 

Summarizing our contributions, we propose 1) a reproducible data collection that includes paired text and speech data in 462 languages, 2) a pipeline and architecture that allows for scaling TTS to an arbitrary number of languages while being highly controllable, 3) a novel loss function that enforces a semantically meaningful structure in the language embedding space, and 4) a procedure that combines meta learning with zero-shot inference enabling the model to synthesize speech in languages for which no data is available.
We evaluate our contributions using objective measures and human evaluation on a set of high-, medium-, and low-resource languages that exhibit a wide range of typological properties. Our code, models, demos, and data are available under an open source license\footnote{\url{https://github.com/DigitalPhonetics/IMS-Toucan}}. 

\section{Proposed Methods}

\subsection{Massively Multilingual Synthesis}

\subsubsection{Data Acquisition and Cleaning}
\begin{table}[t]
    \centering\scriptsize
    \caption{Multilingual datasets and the subset of hours we used. * refers to a new dataset generated from eBible and MMS TTS}
    \begin{tabular}{l r r}
    \toprule
        Dataset & \# Languages & Hours Used  \\ \midrule
        Bible-MMS\textsuperscript{*}                        & 371   & 1230  \\
        Fleurs \cite{conneau2023fleurs}                     & 90    & 377  \\
        Snow Mountain \cite{raju2022snowmountain}           & 15    & 269  \\
        African Voices \cite{ogayo2022africanvoices}        & 11    & 5  \\
        CSS10 \cite{css10}                                  & 8     & 105  \\
        Multilingual LibriSpeech \cite{pratap2020mls}       & 8     & 16,298  \\
        Indian TTS \cite{he2020malayalam}       & 6     & 4.9  \\
        Zambezi Voice \cite{sikasote2023zambezivoice}       & 3     & 0.9  \\
        Living Audio Dataset \cite{braude2019livingaudio}   & 3     & 1.6  \\
        \bottomrule
    \end{tabular} \vspace{-1em}
    \label{tab:data_multi}
\end{table}

To start, we collected a large corpus of publicly available datasets with paired text and speech across various languages, containing over 50,000 hours of data spoken by thousands of speakers. Since such excessive amounts of highly diverse data from many different sources require careful cleaning, we can only use subsets of the full datasets. After a cleaning procedure, which we explain in the following, we end up with around 18,000 hours of data. An overview of the multilingual datasets we used is shown in Table~\ref{tab:data_multi}, and an overview of the monolingual datasets in Table~\ref{tab:data_single}. Most of these datasets are not intended for TTS training and contain noisy recordings, audios with multiple speakers speaking, errors in their labels, and other problems. Therefore, we filtered the audio samples using 1) an open-source speaker diarization system \cite{Bredin23} to retain only excerpts that contain a single speaker, 2) reference-free speech quality metrics to filter out the samples with too much noise, and 3) the loss of our aligner and TTS (see Section~\ref{sec:pipeline}) to find out which samples may have erroneous labels. 

To ensure that the vector space of the language embeddings spans the entirety of possible language embeddings, which we later approximate (see Section~\ref{sec:metalearning}), our pretraining set should be as diverse as possible. Hence, we generate speech using the subsets of the eBible dataset \cite{akerman2023ebible} that are under free licenses as the text input to the MMS TTS models \cite{pratap2023scaling}. By generating 2000 sentences of synthetic speech per language for 371 languages, we substantially increased the linguistic diversity of our data.

\subsubsection{Synthesis Pipeline Design and Training} \label{sec:pipeline}
Our pipeline consists of modular, exchangeable blocks. The general approach is architecture agnostic for most components. We based our implementation on the IMS Toucan toolkit~\cite{lux2023toucan}. First, we converted input texts to a sequence of phonemes. We used eSpeak NG\footnote{\url{https://github.com/espeak-ng/espeak-ng}} for all languages it supports. For the remaining languages, we used transphone \cite{li-etal-2022-zero}, which is a zero-shot phonemizer that provides phoneme annotations for all languages in Glottolog. 
We converted the phonemes into articulatory features (i.e., binary encoded configurations of the vocal tract) 
\cite{lux2022laml}. These articulatory feature sequences were then converted to a mel-spectrogram by a FastSpeech-2-like system \cite{ren2020fastspeech} (50M parameters) that uses FastPitch-style conditioning \cite{lancucki2021fastpitch} on pitch and energy per phoneme, to allow for fine-grained controllability of 
the resulting speech. To obtain the durations needed for this, we made use of a small self-contained aligner \cite{lux2023exact} that was trained with a phoneme recognition objective. To improve details in high frequencies, we used the 
post-net proposed in PortaSpeech \cite{ren2021portaspeech} (40M parameters). This entire synthesis model was conditioned on the outputs of a pretrained speaker-embedding network \cite{speechbrain} to allow for zero-shot voice selection. As a secondary conditioning signal, we enriched the input of the encoder with a language embedding that was learned jointly from a lookup table \cite{lux2022lrms}. 
Everything else was built in a language-agnostic fashion, enabling the zero-shot mechanism described in Section~\ref{sec:metalearning}. The spectrogram that was predicted by the model was then converted into a waveform and upsampled from 16\,kHz to 24\,kHz through the use of a HiFi-GAN vocoder \cite{kong2020hifigan} with additional upsampling steps (14M parameters). Finally, to mitigate the potential of harmful uses, we applied an audio watermark that is robust against modifications \cite{sanroman2024proactive}. 

Training the synthesis model on all data at once, however, is not trivial and fails to converge or has problems with information leakage between language and speaker embeddings. This is likely caused by the 371 synthetic datasets derived from MMS all being single-speaker data, resulting in a high correspondence between language and speaker. To remedy this issue, we employed a training curriculum. First, we trained on a subset of data that consists of only multi-speaker datasets for 40,000 steps. Then, we continued training using all data for a further 120,000 steps with balanced amounts of samples per language per batch. Using eight A6000 GPUs, this training took four days to complete with a combined batch size of 152. Further implementation details can be inferred from our open-source code.

\begin{table}[t]
    \centering\scriptsize
    \caption{Selection of the most important monolingual datasets and the subset of hours we used.}
    \begin{tabular}{l l r}
    \toprule
        Language & Dataset & Hours Used \\ \midrule
        \multirow{3}{*}{English}    
                                    & LibriTTS \cite{zen2019libritts}                           & 236 \\
                                    & HiFi-TTS \cite{bakhturina2021hi}                          & 111 \\
                                    & VCTK \cite{veaux2017superseded}                           & 53 \\
        \midrule
        \multirow{2}{*}{French}     & Blizzard 2023 \cite{perrotin2023blizzard}                 & 29 \\
                                    & SIWIS \cite{yamagishi2019siwis}                           & 11 \\
        \midrule
        \multirow{2}{*}{German}     & HUI-Audio-Corpus \cite{puchtler2021huiaudiocorpusgerman}  & 190 \\
                                    & Thorsten\tablefootnote{\url{https://doi.org/10.5281/zenodo.5525342}} & 34 \\
        \midrule
        Spanish                     & Blizzard 2021 \cite{ling2021blizzard}                     & 6 \\
        Chinese Mandarin            & Aishell-3 \cite{shi2020aishell}                           & 63 \\
        Vietnamese                  & VIVOS \cite{luong2016vivos}                               & 15 \\
        Javanese                    & Javanese ASR \cite{kjartansson2018javanese}               & 59 \\
        Persian                     & ShEMO \cite{nezami2019shemo}                              & 3.1 \\
        Arabic                      & ClArTTS \cite{kulkarni2023clartts}                        & 10 \\
        Amharic                     & ALFFA Amharic \cite{abate2005alffaamharic}                & 2.5 \\
        Swahili                     & ALFFA Swahili \cite{gelas2012alffaswahili}                & 12 \\
        Ukrainian                   & Lada\tablefootnote{\url{https://doi.org/10.5281/zenodo.7396774}}   & 6 \\
        \bottomrule
    \end{tabular} \vspace{-1em}
    \label{tab:data_single}
\end{table}

\subsection{Approximating Synthesis in Unseen Languages}
\subsubsection{Metrics for Language Similarity}
\label{sec:metrics}
For our zero-shot inference mechanism, we need to measure the phonetic distance between languages. Following \cite{wu2021cross}, \cite{li-etal-2022-zero}, and \cite{do2023strategies}, we select three metrics on which we base this distance measure: 1) the distance between nodes over youngest common ancestor, normalized by branch depth in the phylogenetic language tree as not all branches have the same granularity, 2) the distance on the world map, using the ellipsoid distance between language locations according to Glottolog, and 3) the angular similarity of phoneme sets of languages (ASP) based on phonepiece \cite{li2022phone}. Their effectiveness is shown in Section~\ref{sec:lembresults}.

\subsubsection{Language Embedding Space Structure Loss}
\label{sec:less}
We constrain our language embedding space to follow the metrics described in Section~\ref{sec:metrics} by introducing a Language Embedding Space Structure (LESS) loss function $\mathcal{L}_{\mathrm{LESS}}$
which makes the distance between two language embeddings $e(l_1)$ and $e(l_2)$ similar to the average distance between the languages according to the previously discussed metrics with a normalized value range. The exact loss function is given by 
\begin{equation}
    \label{eq:less}
    \mathcal{L}_{\mathrm{LESS}} = \Delta \Biggl( \Delta \Bigl( e\bigl(l_1\bigr) , e\bigl(l_2\bigr) \Bigr) , \cfrac{1}{|M|} \sum_{m \in M}  m\bigl( l_1 , l_2 \bigr)  \Biggr),
\end{equation}
where $\Delta$ denotes the Euclidean distance and $M$ is the set of metrics consisting of the normalized tree distance, the normalized map distance, and the inverse ASP.
In preliminary tests, we found that $\mathcal{L}_{\mathrm{LESS}}$ greatly reduces the chances of the synthesis model diverging by simply adding it to the TTS training loss.

\subsubsection{Meta Learning Unseen Language Representations}
\label{sec:metalearning}
Since our pipeline produces phoneme sequences for any language, to specify the target language, all we need to change within our TTS model is the language embedding. Hence, we can synthesize speech in an unseen language by simply approximating the corresponding language embedding. To achieve this with the limited number of data points we have available (462), we choose to employ a meta-learning technique. Similar to the idea behind Siamese networks \cite{bromley1993signature}, we want to cluster the language embeddings in a latent space, to determine which supervised languages a given language is similar to. We train a three-layer perceptron (96 parameters) as a scoring function that we call Meta Learner ($\mathrm{ML}$) to map pairs of languages, defined by their distance metrics from the set $M$, to approximated language embedding distances. Using the definitions from Section~\ref{sec:less}, we achieve this by optimizing $\mathrm{ML}$ towards fulfilling
\begin{equation}
    \label{eq:metric_learning}
    \Delta \Bigl( e\bigl(l_1\bigr) , e\bigl(l_2\bigr) \Bigr) = \mathrm{ML} \Bigl( m\bigl( l_1 , l_2 \bigr) \text{ for } m \in M \Bigr).
\end{equation}

Using $\mathrm{ML}$ as a learned distance function between languages, we find the $k$ nearest neighbors from our supervised set for an unseen language and average them to approximate the target embedding.
We empirically find the best performance at $5\leq k \leq 25$. Neighbors beyond the minimum are added if their distance falls below a threshold, which we define to be the median distance of the 25th-nearest neighbors across all languages.


\section{Experiments}

In our evaluation of the TTS model, we differentiate between high-resource languages (data is abundant), mid-resource languages (some data is available), and low-resource languages (no data is available). We evaluate two languages for each of these categories and aim for a good spread across the world map and language families. For the high-resource set, we choose English (eng) as a Germanic language and French (fra) as a Romance language to ground the performance of our model in these well-explored settings. For the medium-resource set, we choose Welsh (cym) as a Celtic language and Vietnamese (vie) as an Austroasiatic language, which is also a tonal language. 
For the low-resource language set, we choose Breton (bre), the only Celtic language spoken on the European mainland, and Aymara (aym), an Amerindian isolate language spoken mainly in Peru and Bolivia. We choose to evaluate our approach using real low-resource languages rather than simulating a low-resource scenario by limiting data, despite its
availability, believing this offers a more accurate reflection of the system's potential impact. While this choice narrows our evaluation scope, it aligns our work closely with real-world applications and challenges.

\begin{figure}[t]
    \centering
    \includegraphics[width=.48\textwidth]{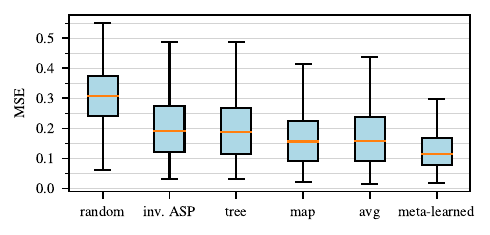}
    \caption{Reconstruction error for approximating the 462 language embeddings from our supervised set using their $k$ nearest neighbors, which are determined either at random, via distance metrics (inverse ASP, tree distance, map distance), their average (avg), or our meta-learned distance function.\vspace{-1em}}
    \label{fig:metalearning}
\end{figure}

\subsection{Language Embedding Approximation}
\label{sec:lembresults}
To evaluate different techniques for approximating language embeddings, we calculated the mean squared error (MSE) between the actual language embeddings from our supervised set and their approximations, obtained by averaging their nearest embeddings selected according to different metrics (see Section~\ref{sec:metalearning}). Figure~\ref{fig:metalearning} demonstrates that our learned metric outperforms any of the individual metrics, as well as their average distance. Furthermore, all metrics perform better than using randomly chosen languages as the nearest neighbors, indicating that they are all effective measures of phonetic language similarity to some extent.

We further conducted a small internal pilot study to determine when the reconstruction error reaches an acceptable level perceptually. We asked participants to listen to simulated zero-shot languages (i.e., using a generated embedding instead of the ground-truth one, despite it being available) in their native language and rate whether the resulting speech sounds natural to them. From this, we find that our proposed metric is the only one which is perceived as sufficiently natural, demonstrating the utility of meta learning for estimating language embeddings. 

Analyzing the selection of nearest neighbors qualitatively shows that the learned metric mostly behaves similar to the map distance, however it pivots to following the tree distance or the ASP if either of them is close to zero or close to one. E.g., Breton is approximated using just five languages: French, Dutch, Hungarian, English and Latin. Notably, Welsh is not used despite being the closest in terms of both map and tree distance, likely due to its higher inverse ASP. Hence the metric seems to be able to generalize to this multi-step policy.

\subsection{Objective Evaluation of the Synthesis}

We computed objective measures for each selected language, with the exception of Aymara, since we lack sufficient quantities of reference speech recordings and appropriate models to measure performance. We based our objective measures on 1000 test samples per language. To evaluate speech intelligibility, we computed the word error rate (WER) between ground truth and automatic transcriptions, obtained by the state-of-the-art automatic speech recognition system Whisper~\cite{radford2023whisper} (version ``large-v3'').
We additionally computed the phoneme error rate (PER) for the phoneme transcripts, which we obtained by applying the phonemizers described in Section~\ref{sec:pipeline} to the ground truth and automatic transcriptions. The PER is less affected by phonetically similar sounding mistakes than the WER, serving as a secondary indicator for intelligibility.
To estimate speech quality, we used WV-MOS, which is a fine-tuned wav2vec2.0 model that aims to predict mean opinion score (MOS) ratings~\cite{andreev2023wvmos}.

\begin{table}[t]
    {
    \setlength{\tabcolsep}{3.5pt}  
    \centering\scriptsize
    \caption{Objective evaluation measures: Word error rate (WER,~$\downarrow$), phoneme error rate (PER,~$\downarrow$), and WV-MOS scores ($\uparrow$). Aymara is excluded for a lack of evaluation resources.}
    \begin{tabular}{lllrr@{\hspace{3pt}}lrr@{\hspace{3pt}}lrr}
        \toprule
        &&& \multicolumn{3}{c}{HighRes} & \multicolumn{3}{c}{MidRes} & \multicolumn{2}{c}{LowRes} \\ \cmidrule{4-5}\cmidrule{7-8}\cmidrule{10-11}
        &&& eng & fra && vie & cym && bre \\ \midrule\midrule
        \parbox[t]{0.1mm}{\multirow{4}{*}{\rotatebox[origin=c]{90}{\hspace{3.5mm}Ours}}}
        && WER    & 0.1 $\pm$ 0.1 & 0.2 $\pm$ 0.2 && 0.3 $\pm$ 0.5 & 0.7 $\pm$ 0.2 && 1.0 $\pm$ 0.3\\
        && PER    & 0.0 $\pm$ 0.0 & 0.0 $\pm$ 0.1 && 0.1 $\pm$ 0.2 & 0.2 $\pm$ 0.1 && 0.7 $\pm$ 0.4\\
        && WV-MOS & 4.4 $\pm$ 0.2 & 3.9 $\pm$ 0.3 && 4.0 $\pm$ 0.3 & 3.6 $\pm$ 0.2 && 4.0 $\pm$ 0.3\\ \midrule
        \parbox[t]{0.1mm}{\multirow{4}{*}{\rotatebox[origin=c]{90}{\hspace{3.5mm}MMS}}}
        && WER    & 0.2 $\pm$ 0.2 & 0.2 $\pm$ 0.2 && 0.3 $\pm$ 0.2 & 0.4 $\pm$ 0.2 && N/A\\
        && PER    & 0.0 $\pm$ 0.1 & 0.0 $\pm$ 0.1 && 0.1 $\pm$ 0.1 & 0.1 $\pm$ 0.1 && N/A\\
        && WV-MOS & 3.9 $\pm$ 0.3 & 4.0 $\pm$ 0.3 && 3.5 $\pm$ 0.4 & 3.6 $\pm$ 0.3 && N/A\\ \midrule
        \parbox[t]{0.1mm}{\multirow{4}{*}{\rotatebox[origin=c]{90}{\hspace{3.5mm}Ref}}}
        && WER    & 0.1 $\pm$ 0.1 & 0.2 $\pm$ 0.2 && 0.1 $\pm$ 0.1 & 0.5 $\pm$ 0.3 && 1.0 $\pm$ 0.4\\
        && PER    & 0.0 $\pm$ 0.0 & 0.0 $\pm$ 0.0 && 0.0 $\pm$ 0.1 & 0.1 $\pm$ 0.1 && 0.5 $\pm$ 0.3\\
        && WV-MOS & 4.1 $\pm$ 0.5 & 3.9 $\pm$ 0.3 && 3.2 $\pm$ 0.4 & 1.9 $\pm$ 1.4 && 2.9 $\pm$ 0.9\\ \bottomrule
    \end{tabular} \vspace{-1em}
    \label{tab:evaluation1}
    }
\end{table}

Table~\ref{tab:evaluation1} shows the objective performance for our system, the MMS system, and a reference obtained by vocoding real-world recordings. Note that the MMS system does not support Breton.
The WER and PER scores show that our and the MMS system are highly intelligible for English, French, and Vietnamese. The high error rates for Welsh and especially Breton (also for the reference) might point towards the low performance of Whisper for these under-resourced languages. Further investigations of the aspect of intelligibility for under-resourced languages remain for future work.
The WV-MOS scores indicate that our system's synthesis quality is on par with or better than MMS. Note that the low scores for the references of some languages are due to the low quality of the reference recordings available.

\subsection{Subjective Evaluation of the Synthesis}

We additionally conducted subjective listening tests, engaging native speakers of the mid- and low-resource languages. These tests were facilitated through an online study using the webMUSHRA framework~\cite{schoeffler2018webmushra}, reaching out to native speakers via research networks and community contacts, ensuring respectful and meaningful engagement with each language community. In this study, we asked the listeners to rate how similar a presented audio sample sounds to someone speaking the respective language as a native speaker on a scale from 1 (``foreign speaker imitating the language without any training'') to 5 (``native speaker of the language'').

\begin{figure}[t]
    \centering
    \includegraphics{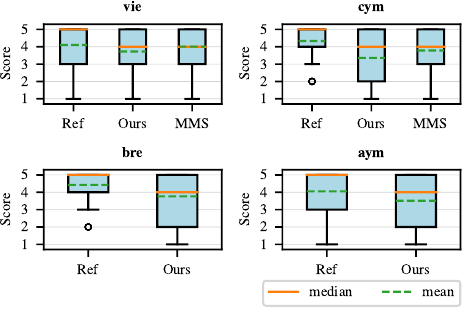}
    \caption{Boxplots for the listening test results. \vspace{-1em}}
    \label{fig:subjective}
\end{figure}

We received 450 ratings from 15 raters for Vietnamese, 390 ratings from 13 raters for Welsh, 200 ratings from 10 raters for Breton, and 180 ratings from 9 raters for Aymara, each evenly spread across all systems.
Note that all participants self-identified as being native speakers of the respective language.
Figure~\ref{fig:subjective} shows boxplots for the obtained listening test scores.
The median score of our system is 4 for all four languages. The MMS system was rated with a median score of 4 as well in the two languages it supports. 
We conducted a Mann-Whitney U test~\cite{rosenberg2017significance} for both Vietnamese and Welsh 
and found no significant difference between the ratings for our system and MMS. This indicates that our system performs on par while supporting nearly seven times as many languages and offering various controllability options.  
Some ratings for the references are lower than expected due to the limited quality of those recordings and differences between language varieties.

\section{Conclusion}
We presented a TTS system that is scalable to an arbitrary number of languages, achieving zero-shot inference on unseen languages through massively multilingual pretraining and a meta-learning approach to approximate language conditioning signals. In this way, we created the first TTS system, which can be used for languages where absolutely no data is available, not even for semi-supervised or transfer learning. The system proves effective across varying resource levels in both objective and subjective evaluation. While the amount of languages covered by the evaluation is a limitation, the linguistic diversity in their selection, as well as the high quality of the ratings by the native speakers, helps ensure the reliability of the results. In the future, it would be interesting to explore if fine-tuning our universal model to language-specific expert models, like MMS, could lead to improvements in those languages.

\section{Ethical Considerations}
Given the risk of misuse of synthetic voice generation, we emphasize that our system is designed for positive applications such as education, accessibility, and cultural preservation. The synthesis is distinguishable from human speech, especially through the use of audio watermarking, as described in Section~\ref{sec:pipeline}.
We further acknowledge the sensitivities associated with using indigenous languages, especially those that communities wish to keep un-documented or limited to specific uses. Our approach involves engaging with community representatives to seek guidance and, where applicable, consent before integrating any language into our system. We are committed to excluding any language from our system upon request from its community, reinforcing our commitment to technology that serves rather than exploits.

\section{Acknowledgments}

 We thank Edwin Banegas-Flores, the voice in the Aymara recordings, and Ruben Hilari-Jilalu, who helped disseminate the Aymara listening test. We thank Anaïs Scornet (\textit{Mignoned ar brezhoneg}) for helping with the Breton test. Lastly, we thank the \textit{Canolfan Bedwyr} center (Bangor University) and the \textit{Mentrau Iaith Cymru} association for their help with the Welsh test.

\newcommand{\BIBdecl}{\setlength{\itemsep}{0.17 em}}
\bibliographystyle{IEEEtran}
\bibliography{bibliography}

\end{document}